\documentclass[conference]{IEEEtran}
\IEEEoverridecommandlockouts
\usepackage{amsmath,amssymb,amsfonts}
\usepackage{graphicx}
\usepackage{textcomp}
\usepackage{xcolor}

{}
{}
{}

\usepackage{adjustbox}
\usepackage{multirow}
\usepackage{color, colortbl}

\definecolor{LightCyan}{rgb}{0.88,1,1}
\definecolor{lightgreen}{rgb}{0.80,1.0,0.80}
\definecolor{lightgray}{rgb}{0.83, 0.83, 0.83}

\usepackage{algorithm}
\usepackage[noend]{algpseudocode}

\algnewcommand\algorithmicforeach{\textbf{for each}}
\algdef{S}[FOR]{ForEach}[1]{\algorithmicforeach\ #1\ \algorithmicdo}

\usepackage[font=footnotesize]{caption}
\usepackage{subfig}
\usepackage{booktabs} 
\usepackage{multirow}
\usepackage{diagbox}
\usepackage[compress]{cite}

\renewcommand{\baselinestretch}{0.93}
\def\BibTeX{{\rm B\kern-.05em{\sc i\kern-.025em b}\kern-.08em
    T\kern-.1667em\lower.7ex\hbox{E}\kern-.125emX}}

\begin{document}
\bstctlcite{IEEEexample:BSTcontrol}


\title{Conditional Classification: A Solution for Computational Energy Reduction
}

\author{\IEEEauthorblockN{Ali Mirzaeian$^*$, Sai Manoj$^*$, Ashkan Vakil$^*$, Houman Homayoun$^\dagger$, Avesta Sasan$^*$ }
$^*$Department of ECE, George Mason University, e-mail: \{amirzaei,  spudukot,avakil, asasan\}@gmu.edu\\
$^\dagger$Department of ECE, University of California, Davis, e-mail: hhomayoun@ucdavis.edu}

\maketitle
\begin{abstract}
Deep convolutional neural networks have shown high efficiency in computer visions and other applications. However, with the increase in the depth of the networks, the computational 
complexity is growing exponentially. 
In this paper, we propose a novel solution to reduce the computational complexity of convolutional neural network models used for many class image classification. Our proposed technique breaks the classification task into two steps: 1) coarse-grain classification, in which the input samples are classified among a set of hyper-classes, 2) fine-grain classification, in which the final labels are predicted among those hyper-classes detected at the first step. We illustrate that our proposed classifier can reach the level of accuracy reported by the best in class classification models with less computational complexity (Flop Count) by only activating parts of the model that are needed for the image classification.
\end{abstract}

\vspace{2mm}

\section{Introduction}\label{sec:introduction}
The research and development of Deep Neural Networks (DNNs) combined with the availability of massively parallel processing units for training and executing them, have significantly improved their applicability, performance, modeling capability, and accuracy. Many of the recent publications and products affirm that the state-of-the-art DNN solutions achieves superior accuracy in a wide range of applications compared to the outcome of the same task that is performed by other na{\"i}ve techniques \cite{sayadi20192smart, vakil2020lasca, azar2020nngsat, vakil2020learning, he2016deep}. DNN models are specially powerful for solving problems that either have no closed-form solution or are too complex for developing a programmable solution. 

The trend of development, deployment, and usage of DNN is energized by the rapid development of massively parallel processing hardware (and their supporting software) such as  Graphical Processing Unit (GPU) \cite{sanders2010cuda}, Tensor Processing Units (TPUs) \cite{abadi2016tensorflow}, Field Programmable Gate Arrays (FPGAs), Neural Processing Units (NPUs) \cite{chen2014diannao, chen2014dadiannao, du2015shidiannao, chen2016eyeriss, mirzaeian2019tcd, nesta, daneshtalab2020hardware, Faraji_ISCAS_2020, 9116473, 10.1145/3427377}, and many-core solutions for parallel processing of these complex, yet massively parallelizable models.  

The ability to train and execute deeper models, in turn, has resulted in significant improvement in the modeling capability and accuracy of CNNs, a trend that could be tracked from early CNN solutions such as 5-layer Lenet-5 \cite{lecun2015lenet} for handwritten digit detection, to 
much deeper, complex, and fairly sophisticated 152 layer ResNet-152 \cite{he2016deep} used for 1000-class image classification with an accuracy that significantly surpasses that of human capability. Generally, going deeper (or wider) in CNNs improves their accuracy at the expense of increased computational complexity. Increasing the model complexity reduces the range of hardware that could execute the model within an acceptable time and could justify the extra energy consumed for executing a deeper yet more accurate model \cite{neshatpour2018icnn}. Hence, many researchers in the past few years have visited the problem of reducing the computational complexity of CNNs \cite{xiao2014error, neshatpour2018icnn, Heidari-etal-2020-ELMO, srivastava2013discriminative, deng2014large, liu2013probabilistic, Heidari-etal-2020-Social_bots} for widening their use and applicability.  

In this paper, we propose an efficient solution to reduce the computational complexity of CNNs used for many-class image classification.  Our proposed model breaks the classification task into two stages of 1) Clustering, and 2) Conditional Classification. 
More precisely we transform a difficult $K$-class classification problem into a $K_1$-group clustering and in which each cluster contains $C_i$ classes i.e., $K = \sum_{i=1}^{K_1}{C_i}$. The $K_1$ group (a.k.a Hyper-Class) clustering problem is solved by a convolutional encoder (first-stage of our proposed model). In this model, each Hyper-class is composed of a set of classes with shared features that are closely related to one another. The decision of which classes are grouped into the same cluster is made by applying the spectral clustering algorithm \cite{ng2002spectral}  on the similarity matrix obtained from the K-Nearest Neighbour algorithm (KNN) \cite{shi2000normalized} on the latent spaces corresponding to the input samples. After validating the membership of an input image to a cluster, the output of the encoder is pushed to a corresponding class classifier that is specifically tuned for the classification of that hyper-class. By knowing the hyper-plane (cluster-plane), the complexity of detecting the exact class is reduced as we can train and use a smaller CNN when classification space (the number of classes) is reduced.

To generalize the solution, we formulate a systematic transformation flow for converting the state of the art CNNs (original model) into a 2-stage Clustering-Classification model with significantly reduced computational complexity and negligible impact on the classification accuracy of the overall classifier. 

\section{Related Work}


 The problem of model complexity reduction has also been visited by many scholars. A group of related previous studies has addressed the problem of reducing the \textit{average-case} computational complexity by breaking the CNN models into multiple stages and giving the option of an early exit using deploying mid-model classifiers \cite{neshatpour2018icnn, panda2016conditional, teerapittayanon2016branchynet}. For example, in \cite{neshatpour2018icnn, neshatpour2019icnn, neshatpour2018design} it was shown that the average computational complexity of the model (over many input samples) could be reduced by breaking a large CNN model into a set of smaller CNNs that are executed sequentially. In this model, each smaller CNN (uCNN) can classify and terminate the classification if an identified class has reached a desired and predefined confidence threshold. Similarly, in \cite{panda2016conditional}, a Conditional Deep Learning Network (CDLN) is proposed in which, fully connected (FC) layers are added to the intermediate layers to produce early classification results. The forward pass of CDLN starts with the first layer and monitors the confidence to decide whether a sample can be classified early, skipping the computation in the proceeding layers. While CDLN only uses FC layers at each exit point, BranchyNet \cite{teerapittayanon2016branchynet} proposes using additional CONV layers at each exit point (branch) to enhance the performance. Unfortunately, this group of solutions suffers from 2 general problems: 1) although, they reduce the average-case computational complexity, their worst-case complexity (when all uCNN or additional FC and CONV layers are executed) is worse than the state of the art's non-branchable solutions. 2) Introducing many additional Fully Connected (FC) layers makes them suffer from a parameter-size explosion as FC layers require a far large number of parameters than CONV layers, worsening their memory footprint. Our proposed solution addresses the shortcomings of these models by making the execution time uniform across different input samples, keeping the FC layer memory footprint in check, while reducing the computational complexity of the model.

 On the other hand, utilizing hierarchical structures for training and inference phase of Convolutional Neural Networks for improving their classification accuracy has been studied and addressed in many previous works \cite{xiao2014error, neshatpour2018icnn, srivastava2013discriminative, deng2014large, liu2013probabilistic}. However, the focus of most of these studies was on improving the accuracy rather than addressing its complexity problem. Notably, in some of these studies, it is shown that employing hierarchical structures could degrade the model's efficiency. For example, in  \cite{yan2015hd}, the authors reported an increase in both memory footprint and classification delay (computational complexity) as noticeable side effects of deploying hierarchical classification for improving the model's accuracy. Similar to this group of studies, we explore the hierarchical staging of CNN models, but with a different design objective: We propose a systematic solution for converting a CNN model into a hierarchical 2-stage model that reduces the computational complexity and model's memory footprint with negligible impact on its accuracy. 
 



\begin{figure}[!ht]
    \centering
   \includegraphics[width=0.85\columnwidth]{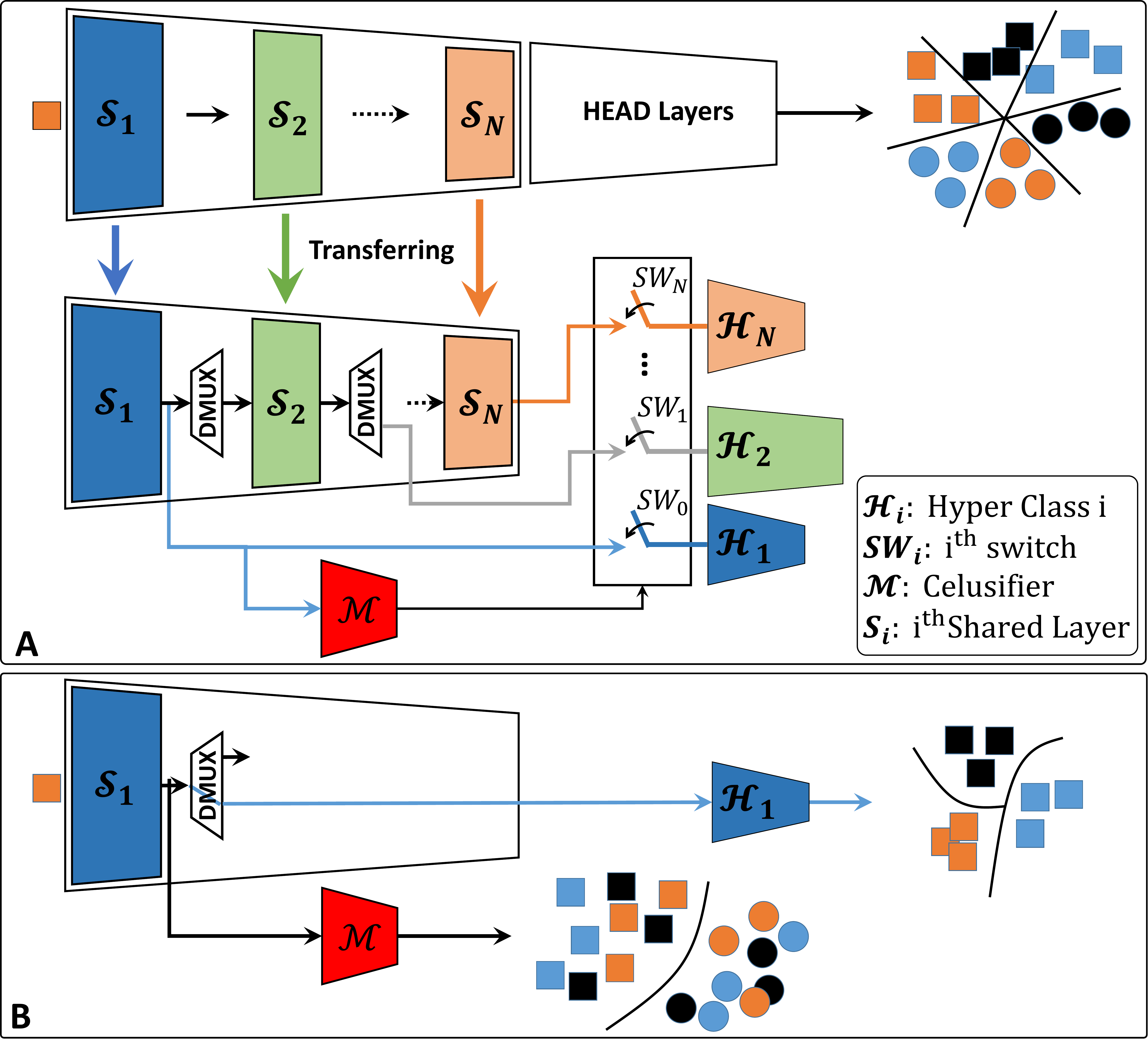}
    \caption{A) Employing the proposed method for training a hierarchical architecture. B) An example of functionality the hierarchical cluster method for classifying an input sample shown with orange square.  SW$_0$ to SW$_N$ are N switches that controls which path should connect, and the Hyper-class1 to Hyper-classN, shown with $\mathcal{H}_1$ to $\mathcal{H}_N$, are the clusters in which trained separately and attached together. }
    \label{fig:overview}
\end{figure}

\section{Proposed Method}
A CNN model is composed of several Convolution (CONV) layers and usually one or more Fully Connected (FC) layer for final classification. Each CONV layer extracts a set of features from its input feature map (ifmap) and generates a more discriminative output feature map (ofmap). The ofmap of a CONV layer is the ifmap to its proceeding CONV layer. The CONV layers close to image input will become specialized in extracting generic (class-independent) features. But, as we move deeper into the CNN, the CONV layers extract more abstract (higher-level representation) features of the input image from their ifmap. The CONV layers close to the output (softmax layer) become specialized in extracting the most abstract and class-specific features. This allows the last layer (i.e. FC and softmax layer) to identify and assign a probability to each class based on the activation of neurons in the last CONV layer. In short, earlier CONV layers extract low-level features needed for the classification of all input images, while the later CONV layers are specialized for extracting abstract features for the classification of specific classes. 

Motivated by this view of CONV layers' functionality, we present a simple yet efficient and systematic solution to re-architect the state-of-the-art CNN models into a hierarchical CNN model such that any given input image activates only the part of the model that is needed for the classification. Our proposed (target) model architecture, as illustrated in Fig. \ref{fig:overview}-A bottom, is composed of three main modules: (1) $\mathcal{S}$: $\mathcal{S}$hared Clustering layer(s), (2) $\mathcal{M}$: $\mathcal{M}$id cluster classifier(s) (a.k.a. clustifier), and (3) $\mathcal{H}$: a set of $\mathcal{H}$yper-class specific micro CNN models. The $\mathcal{S}$ layer is used to extract low-level features from an input image, while the $\mathcal{M}$ layer is used for classifying the input image into one of $K_1$ clusters (hyper-classes). Based on the result of clustifier $\mathcal{M}$, the associated cluster-specific model $\mathcal{H}$ is activated to classify the image to one of its possible $K_2$ classes. Considering that $K_1<<K$ and $K_2<<K$, clustering and classification could be performed by a much shallower CNN. Also, note that we can have clusters of different sizes. In this model, we divide the $K$-class clustering problem into $K_1$ clustering problem, each containing $C_i$ classes such that $K = \sum_{i=1}^{K_1}{C_i}$, while still honoring the $K_1<<K$ and for each $i$, $C_i<<K$. Finally note that, as illustrated in Fig. \ref{fig:overview}-A, by using additional $\mathcal{S}$ and $\mathcal{M}$ layers, we can hierarchically break a large cluster into smaller clusters and use a dedicated $\mathcal{H}$ for each of the smaller clusters, while allowing many of the clusters to share a larger set of shared ($\mathcal{S}$) CONV layers. 


\subsection{Proposed Architecture}

To build our proposed model, we designed (1) a mechanism to break and translate a state of the art CNN into a trainable 3-stage model that preserves the model accuracy, and (2) developed an effective solution for clustering classes with shared features into the same cluster. Details of our systematic solutions for constructing the model and its training are discussed next.

\begin{figure*}[hbt!]
    \centering
    \includegraphics[width=1.6\columnwidth]{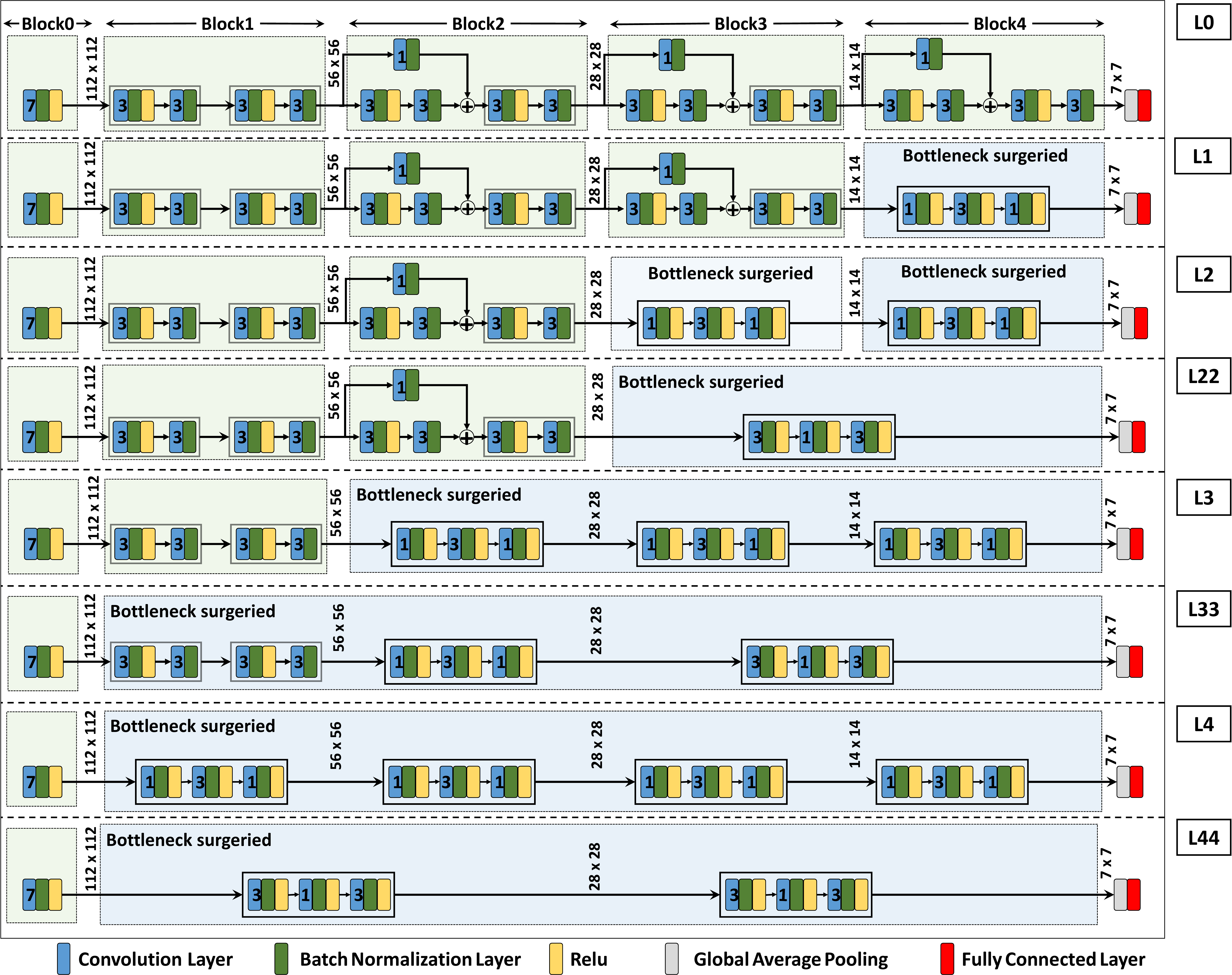}
    \caption{Generated micro CNNs after applying Bottleneck surgery for Resnet18. The green blocks in the first section of the table indicate CONV layers borrowed (not changed) from the original Resnet18 (and froze during the training). The blue blocks are the sections that are replaced with a bottleneck or bottleneck-compression blocks and are trained.}
    \label{bts}
\end{figure*}

\textbf{Shared Classifier $\mathcal{S}$:} At this stage by 
sweeping all the design spaces, we specify which layers of the original model lay in the shared part of the hyper classes. For example, Fig. \ref{fig:overview} shows a scenario that for the hyper class one, $\mathcal{H}_1$, only the parameters of the first layer of the original model has been transferred, whereas, for the hyper class two, $\mathcal{H}_2$, the parameters of the first two layers of the original model has been employed. A detailed case study on Resnet18 has been shown in Fig. \ref{bts}.


\textbf{Mid Clustifier $\mathcal{M}$:} The implementation of clustifier $\mathcal{M}$ is more involved, as the performance of clustifier $\mathcal{M}$ significantly
impacts the accuracy of the overall solution. For a given input $X$, if $\mathcal{M}(X)$ actives an incorrect hyper-class classifier, the input is miss-classified. To improve the accuracy of the proposed solution, we propose a
confidence-thresholding mechanism in which the clustifier $\mathcal{M}$ could activate a minimum set of hyper-class classifiers, such that the confidence of the clustifier $\mathcal{M}$ in the inclusion of the correct hyper-class classifier in the minimum activation set is above a given threshold.  

To achieve this objective, the clustifier $\mathcal{M}$ considers the cluster probabilities (confidence) suggested by the clustifier along with the data in the confusion matrix (CM) of the clustifier $\mathcal{M}$ to activate the related hyper-classes for each input sample $X$.  The confidence of the clustifier is the probability suggested by the softmax layer of the Clustifier $\mathcal{M}$ for the input label. The confusion matrix of the clustifier is a two-dimensional table that contains the confusion score of each class with other classes and is obtained by benchmarking the clustifier $\mathcal{M}$ using a set (i.e. test set) of labeled inputs. In this paper, $\mathcal{P}_{ij}$ is the value of $i^{th}$ unit of the confusion matrix when $j^{th}$ label is predicted. We also use the notation $CC_i(X)$ to refer to the $i^{th}$ highest score class that is confused with the class of input $X$ as suggested by the confusion matrix, where $i$ determine the ranking of confused class in the matrix i.e., $i=1$ represents the class that is mostly confused with the class of $X$.  

To increase the likelihood of including the correct hyper-class classifier in the activation set, we first define a confidence threshold $\tau_{CS}$ (i.e. 90\%) and a variable $CS$ for holding the confidence summation results which is initially set the highest cluster probability suggested by $\mathcal{M}(X)$. If the clustifier's confidence (suggested probability) for the selected hyper-class is below the confidence threshold, we refer to the confusion matrix of the clustifier $\mathcal{M}(X)$, and select the hyper-class $CC_i(X)$ (i.e. $i=1$, for the class most confused with the selected class). Then we find the suggested confidence of the selected hyper-class from $\mathcal{M}(X)$, and add the suggested confidence to the $CS$. This process is repeated until the $CS > \tau_{CS}$. The exit condition is expressed in Eq. \eqref{eq:cs}. At this point, the clustifer activates all selected classes in the set contributing to the $CS$. This procedure is outlined in the Alg. \ref{policy_algorithm}.   

\begin{equation}
\label{eq:cs}
   \underset{K}{CS = \operatorname{argmin}} 
 \sum_{i=1}^{i=k}(\mathcal{V}(\mathcal{A}(i))) > \tau_{CS}
\end{equation}  

\begin{algorithm}[hbt!]
	\caption{Hyper-class activation policy}
	\label{policy_algorithm}
	\begin{algorithmic}
		\footnotesize
		
		\Procedure{Activator}{Clustifier $M$, Input $X$, Confidence threshold $\tau_{CS}$, Hyperclass pool $HPool$, Confusion Matric $CM$}
    	\State V = $M(X)$
		\For{($l=1$; size of $HPool$ ; $l++$)}
    		\If{$l==1$}
    		    \State index = argmax(V)
    		    \If{V[index] $ > $ $\tau_{CS}$}
    		       \State activate HPool[index]
    		       \State $Exit.$
    		    \EndIf
		    \EndIf
		    
		    \State actives = argNmax(N=$l$, CM[:,index])
		    \State temp = Nmax(N=$l$, CM[:,index])
		    \State cmVal = temp/sum(temp)
		    
    		\If{sum(V[actives]) $ > \tau_{CS}$ }
		        \State activate  (HPool[actives]*cmVal)
		        \State $Exit.$
		    \EndIf		    
		\EndFor
		\EndProcedure 
		\normalsize
	\end{algorithmic}
\end{algorithm}

\begin{figure}[hbt!]
    \centering
    \includegraphics[width=0.99\columnwidth]{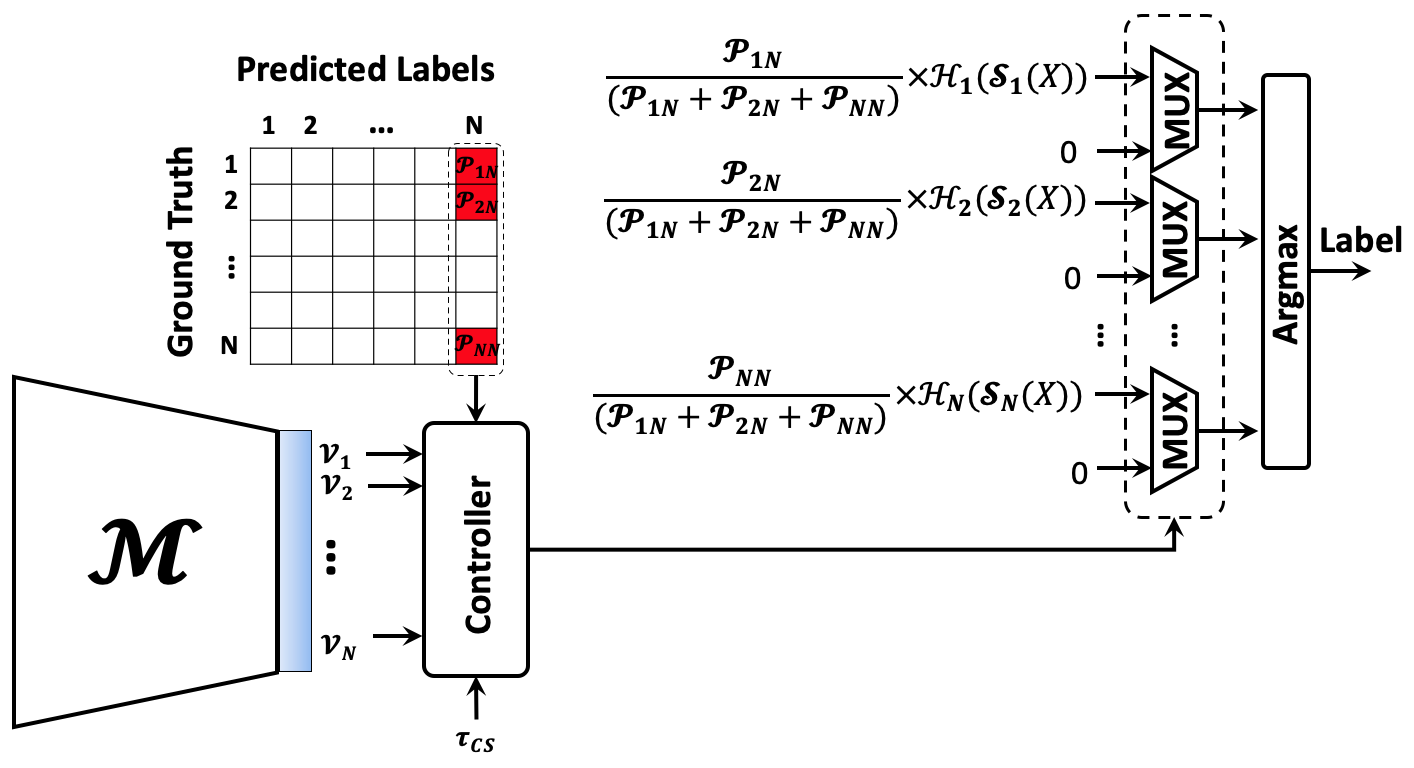}
    \caption{A example of Hyper-class Selection policy. X is the sample input, $\mathcal{V}_i$ is the output of the clustifier's softmax layer, $\mathcal{S}_i(X)$ is the output of the shared layer related to $i^{th}$ hyper class, $\mathcal{P}_{ij}$ is the value of $i^{th}$ unit of the confusion matrix when $j^{th}$ label is predicted and finally $\mathcal{H}_i(\mathcal{S}_i(X)$) is the dedicated part of the $i^{th}$ hyper class. }
    \label{fig:selector-example}
\end{figure}

Fig. \ref{fig:selector-example} shows an example of this algorithm when three hyper-classes are activated. In this example the clustifier has predicted the label N for the input sample $X$, however, its confidence, $\mathcal{V}_N$ , doesn't pass the defined threshold $\tau_{CS}$. So the $CC_1(X)$ and $CC_2(X)$ that respectively have probability $\mathcal{V}_{1}$ and $\mathcal{V}_{2}$ (as suggested by $\mathcal{H}(x)$) are added to activation set. 

The next challenge for training a clustifier is identifying which classes could be grouped to improve the accuracy of the clustifier. We propose that grouping similar  classes in a cluster is an efficient solution for achieving high accuracy while keeping the computational and model complexity of the clustifier in check. Note that this approach, improves the accuracy of the mid-clustifier at the expense of posing a harder the task to the hyper-class classifier. Nevertheless, because the hyper-class classifier is a deeper network than the mid-classifier, it should be more capable in descriminating between classes that are grouped in the same cluster for higher similarity. To achieve our objective of grouping similar classes in the same cluster, we employed the unnormalized spectral clustering introduced in   \cite{ng2002spectral}, \cite{shi2000normalized}. Note that the cluster sizes in this approach are not uniform, suggesting that the size of the hyper-class classifiers could also be different. Our implementation of spectral clustering is discussed next:

Given a set of points $S = \{S_1,..., S_n$\} $\in$ $\mathbb{R}^l$, they can be clustered into $k$ coarse classes following the algorithm \ref{Extractor}. First step of using spectral clustering is to define a similarity matrix between different classes. For obtaining the similarity matrix, we first obtain the probability of each class on a (labeled) evaluation set. Further, we compute the average probability vector of each class across all input images available for that class in the evaluation set. This vector of probabilities is known as indicator vector, denoted by $V_i$, and computed using Eq. \eqref{eq:cc}. 

\begin{equation}
    \label{eq:cc}
        V_i = (1/M)\sum_{j=1}^M(prob[j]*(i==gt_j))
\end{equation}

In this equation, $gt_j$ is the ground truth label for image $j$, and the prob[j] is the vector of probabilities generated for image $j$. The next step is to apply the K-Nearest Neighbour(KNN) clustering on the indicator vectors to build a similarity matrix. Connectivity parameter of KNN algorithm indicates the number of the nearest neighbors, has been set to the first value at the range [1, N] which leads to a connected graph and that is because the spectral clustering algorithm, has its best functionality when the similarity matrix represents a connected graph. The similarity matrix feeds to the unnormalized Spectral Clustering algorithm and using the eigengap heuristic \cite{von2007tutorial} the number of suitable coarse classes are selected. For example, when following algorithm \ref{Extractor}, the obtained number of hyper-classes for CIFAR100 dataset is 6, and the number of members at each hyper-class are 9, 28, 23, 15, 14, 11 regards to c0, c1, c2, c3, c4, c5, respectively (see Table. \ref{resnet18} in the result section).

\begin{algorithm}[hbt!]
	\caption{Cluster Membership Assigment}
	\label{Extractor}
	\begin{algorithmic}
		
		\footnotesize
		\Procedure{Extractore}{$S_1$, $S_2$, ..., $S_n$}
		\State 1) Constructing similarity matrix A using K-Nearest Neighbor(KNN):
        \State ~~~A = KNN($S_1$, $S_2$, ..., $S_n$).
    	\State{2) Define Degree matrix D:}
    	\State ~~~$D_{ii} = \sum_jA_{ij},\,\,\,\,D_{ij} = 0 \,\,\,\,if i\neq j $.
    	\State{3) Constructing unnormalized Laplacian matrix L:}
    	\State ~~~$L = D-A$.
    	\State{4) solving the generalized eigenproblem:}
    	\State ~~~$Lx=\lambda Du$.
    	\State{5) $X = [x_1 x_2 ... x_k] \in \mathbb{R}^{n\times k}$ related to the lowest k eigenvalues of $L$}
    	\State{6) Construct the matrix Y as:}
    	\State ~~~$Y_{ij} = X_{ij}/(\sum_jX^2_{ij})^{1/2}$.
    	\State{7) Apply K-means on each row of Y as a data point in $\mathbb{R}^K$}. 
    	\State{8) datapoint $S_i$ $\in$ cluster j, if and only if $Y_i$ $\in$ j.}
    	
		\EndProcedure\\ 
		\normalsize
	\end{algorithmic}

\end{algorithm} 

\begin{table*}
    \centering
    \scriptsize
    \caption{L1, L2, L3, L4, L22, L33, L44 are some of the compressed micro CNNs that could be generated from the application of our proposed model compression flow on Resnet18.  The top table presents the top1 accuracy for different clusters when different compressed CNN configurations are used for classification, while the bottom section captures the associated flop count (computational complexity). Units marked with * shows one of the possible mapping of each cluster to the corresponding structure, accordingly the selected configuration is \{C0:L44, C1:L1, C2:L1, C3:L1, C4:L44, C5:L44, Clustifier:L44\} in which X:Y means cluster X is mapped to the structure Y.}
    \label{resnet18}
	\begin{tabular}{|c|c|c|c|c|c|c|c|c|c|}
		\hline 
		\rowcolor{lightgray}
		\multicolumn{10}{|c|}{\textbf{Top1 Accuracy}}\tabularnewline
		\hline 
		\textbf{L-Name} & \textbf{O-Size} & \textbf{L0} & \textbf{L1} & \textbf{L2} & \textbf{L22} & \textbf{L3} & \textbf{L33} & \textbf{L4} & \textbf{L44}\tabularnewline 
		\hline
		\textbf{C0}             & 9      & 0.83	    &0.9                &0.898	&0.892  &0.869	 &0.869   &0.843	&\textbf{0.843}* \tabularnewline    
		\hline                                                                                   
		\textbf{C1}             & 28     & 0.671	&\textbf{0.769}*	&0.726	&0.705  &0.687	 &0.666   &0.660	&0.601\tabularnewline    
		\hline                                                                                   
		\textbf{C2}             & 23     &  0.623	&\textbf{0.748}*    &0.716  &0.691  &0.708   &0.689   &0.673   	&0.667\tabularnewline    
		\hline                                                                                   
		\textbf{C3}             & 15     & 0.800	&\textbf{0.810}*    &0.791  &0.755  &0.75	 &0.707   &0.753	&0.650\tabularnewline    
		\hline                                                                                   
		\textbf{C4}             & 14     &  0.832	&0.921	            &0.886  &0.900  &0.876	 &0.894   &0.871	&\textbf{0.841}* \tabularnewline    
		\hline                                                                                   
		\textbf{C5}             & 11     & 0.757	&0.855	            &0.861  &0.853  &0.845	 &0.816   &0.841   	&\textbf{0.832}*\tabularnewline    
		\hline                                                                                   
		\textbf{Clustifier}     & 6      & 0.869	&0.953	            &0.876	&0.957  &0.941	 &0.919   &0.931	&\textbf{0.923}* \tabularnewline    
		\hline  
		\rowcolor{lightgray}
		\multicolumn{10}{|c|}{\textbf{Computational Complexity (Flops) Reduction}}\tabularnewline
		\hline
		\textbf{C0}             & 9      & 1823527936 (0\%)	& $\sim$ 14.7\%	&$\sim$28.7\%	&$\sim$42.8\%	&$\sim$60.5\%	&$\sim$39.7\%		&$\sim$53.8\%	&$\sim$79.3\%\tabularnewline 
		\hline
		\textbf{C1}             & 28     & 1823531008 (0\%)	&$\sim$14.7\%	&$\sim$28.7\%	&$\sim$42.8\%	&$\sim$60.5\%	&$\sim$39.7\%	&$\sim$53.8\%	&$\sim$79.3\% \tabularnewline   
		\hline                                                                                               
		\textbf{C2}             & 23     & 1823550464 (0\%)	&$\sim$14.7\%	&$\sim$28.7\%	&$\sim$42.8\%	&$\sim$60.5\%	&$\sim$39.7\%	&$\sim$53.8\%	&$\sim$79.3\%\tabularnewline    
		\hline                                                                                               
		\textbf{C3}             & 15     & 1823545344 (0\%)	&$\sim$14.7\%	&$\sim$28.7\%	&$\sim$42.8\%	&$\sim$60.5\%	&$\sim$39.7\%	&$\sim$53.8\%	&$\sim$79.3\%\tabularnewline    
		\hline                                                                                               
		\textbf{C4}             & 14     & 1823537152 (0\%)	&$\sim$14.7\%	&$\sim$28.7\%	&$\sim$42.8\%	&$\sim$60.5\%	&$\sim$39.7\%	&$\sim$53.8\%	&$\sim$79.3\% \tabularnewline    
		\hline                                                                                               
		\textbf{C5}             & 11     & 1823536128 (0\%)	&$\sim$14.7\%	&$\sim$28.7\%	&$\sim$42.8\%	&$\sim$60.5\%	&$\sim$39.7\%	&$\sim$53.8\%	&$\sim$79.3\%\tabularnewline    
		\hline                                                                                               
		\textbf{Clustifier}     & 6      & 1823533056 (0\%)	&$\sim$14.7\%	&$\sim$28.7\%	&$\sim$42.8\%	&$\sim$60.5\%	&$\sim$39.7\%	&$\sim$53.8\%	&$\sim$79.3\% \tabularnewline 
		\hline  
	\end{tabular}

\end{table*}

\textbf{Hyper-Class classifier $\mathcal{H}$}: The hyper-class classifiers are micro CNNs that are trained from scratch and become specialized in classifying each cluster. Considering that the size of clusters may be different, the size of the hyper-class classifiers may also vary. To design the hyper class classifiers we need to solve two issues: 1) considering that more than one $\mathcal{H}$ may be activated at a time, we need to find a solution to select or sort the suggested classes by different $\mathcal{H}$s; 2) we need a mechanism to transform the non-shared portion of the original CNN to these smaller and hyper-class specific micro CNNs. Each of these is discussed next:

For solving the first problem of simultaneous activation of multuple $\mathcal{H}$, we propose sorting the weighted confidence of $\mathcal{H}$ classifiers' prediction and choose the top (i.e. top 1 or top 5) as the prediction of the overall model. We propose using the scores obtained from the confusion matrix (which was used for activation of hyper-class classifiers) to compute the weighted class probabilities and then sort the weighted probabilities to determine the top 1 or top 5 classes.  The Eq. \eqref{eq:fusion} illustrates how the class probabilities are weighted for the example given in Fig. \ref{fig:selector-example}.

\begin{equation}
  \label{eq:fusion}
  \underset{f\in {1, 2, N}}{\operatorname{argmax}} 
   ({\mathcal{P}}_{fN}/( {\mathcal{P}}_{1N}+{\mathcal{P}}_{2N}+{\mathcal{P}}_{NN} ) \times \mathcal{H}_f(X))   
\end{equation}

The next problem is designing the micro-CNNs that act as hyper-class classifiers. For this purpose, we propose a solution to automate the transformation of non-shared layers of the original model to micro CNN models. For this purpose, we propose reducing the size of non-shared CONV layers by replacing some of the CONV layers with a combination of two CONV layer configurations $ \left(\begin{array}{c} 1\times1,x \\ 3\times3,y\\ 1\times1,z \end{array}\right)$ and $ \left(\begin{array}{c} 3\times3,x \\ 1\times1,y\\ 3\times3,z \end{array}\right)$, in which $f\times f, x$ shows a kernel size $f \times f$ with $x$ channels. The first block is known as a bottleneck block, and we denote the second block as bottleneck-compression block. 

\begin{figure}[hbt!]
    \centering
    \includegraphics[width=0.8\columnwidth]{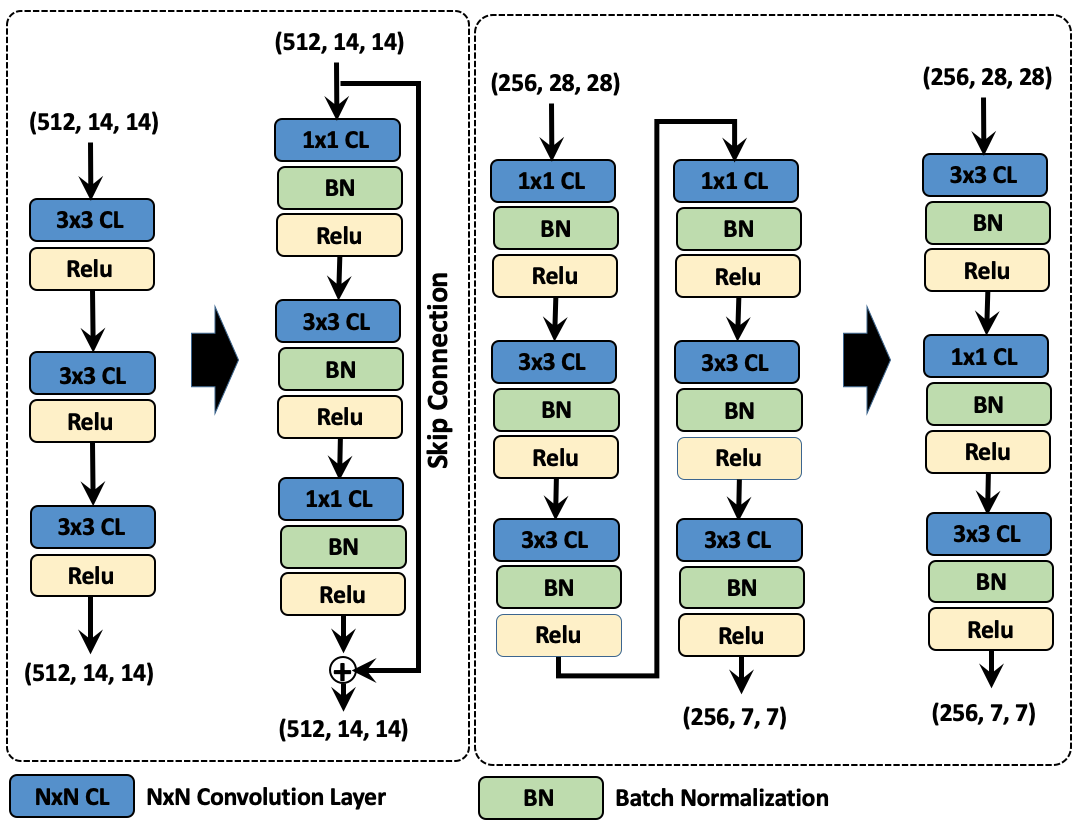}
    \caption{Left: compression flow for a hypothetical target block. Right: replacing two consecutive bottle-neck with a bottle-neck compression block.  At the configuration (X, Y, Z), X is the channel number, Y and Z are width and height of the image shape.}
    \label{model_compression_flow}
\end{figure}

Our model compression flow is as following: 1)  Starting from the last CONV layer of the original model, we identify target blocks that could be replaced with bottleneck layers. Let's assume the ifmap to the first CONV layer an identified block is $x_1, y_1, c_1$ and the ofmap of the last CONV layer in the identified block is $x_2, y_2, c_2$, in which $x$ and $y$ are the width and height of each channel, and $c$ is the number of channels. In this case the targeted block could be replaced by a bottleneck block if $x_1 = x_2 \,\, and \,\, y_1 = y_2$ or $x_1/2 = x_2 \,\, and \,\, y_1/2 = y_2$. In the first case, the stride of the bottleneck block is set to 1, and in the second case, the stride is set to 2. In addition, for each targeted block if $c_1=c_2$ an skip connection is added.  The compression could be pushed further by identifying two consecutive bottleneck blocks and replacing it with a bottle-neck compression block. This translation process is illustrated in Fig. \ref{model_compression_flow}. Depending on how many bottlenecks or bottleneck-compression blocks are inserted, we can have a wide range of compressed CNNs.

\section{Experimental Results}
\subsection{Evaluating the Model Compression Solution} \label{result_compression}
We first illustrate the effectiveness of our propose compression process in terms of its impact on model complexity and accuracy. For this purpose, we apply our solution to compress the Resnet18. We also used the algorithm \ref{Extractor} to divide the CIFAR100 data set into different clusters. The algorithm suggested 6 clusters with 9, 28, 23, 15, 14, 11 classes in each hyper-class. These hyper classes are respectively denoted as c0, c1, c2, c3, c4, c5. 

The first section of the Table. \ref{resnet18} captures some of the possible configurations, showed in Fig. \ref{bts}, from the application of bottleneck and bottleneck-compression blocks on Resent18. As illustrated, the compression solution generates a wide range of compressed micro CNN. The second section of the table captures the accuracy of the compressed network for each cluster and each compressed network configuration, while the third section captures the reduction in the complexity for each compressed model (compared to the L0 -original- case). As illustrated, the compressed networks are still able to achieve very high accuracy with a significant reduction (up to 79\%) in their computational complexity.

\subsection{Accuracy and Complexity Evaluation}

In section \ref{result_compression} only the accuracy of a model composed of the shared $\mathcal{S}$ CONV layer (green blocks in Fig. \ref{bts}) and hyper-class specific compressed layers $\mathcal{H}$s (blue blocks in Fig. \ref{bts}) was evaluated. However, the overall accuracy of the model is also impacted by the accuracy of the Mix Clustifier $\mathcal{M}$ and the combined accuracy of selected Hyper-class classifiers (i.e. $\mathcal{H}s)$. To evaluate the overall accuracy of the resulted model we selected the following configurations for building the hyper-class classifiers for each of the 6 clusters that we previously identified: \{C0:L44, C1:L1, C2:L1, C3:L1, C4:L44, C5:L44, Clustifier:L44\}. These configurations are highlighted with an * in Table \ref{resnet18}.  We report the accuracy and complexity result of our model that we evaluated for 10,000 images of CIFAR100 in our test set. 

\begin{figure}[hbt!]

    \centering
    \includegraphics[width=0.7\columnwidth]{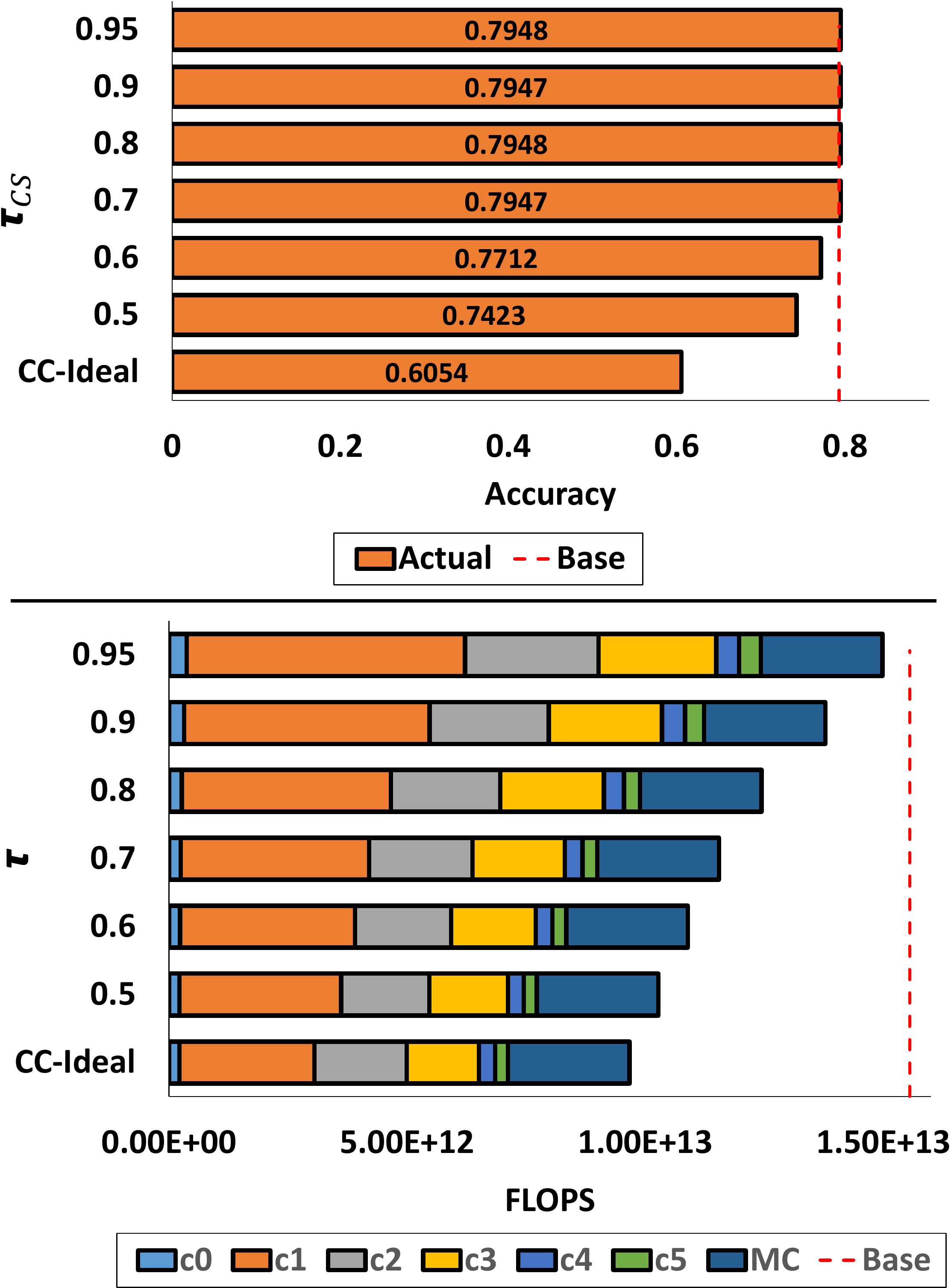}
    \caption{ Depicting the impact of changing $\tau_{CS}$ on accuracy and computational complexity (CC). $\tau_{CS}$ indicates the confusion sum threshold and the CC metric has been calculated in FLOPS. }
    \label{fig:metrics}
    
\end{figure}

Table \ref{activated} captures the number of activated Hyper-Classes (HC) when the confidence summation, $\tau_{CS}$, varies in the range (0.5, 0.95). As illustrated in Table \ref{activated}, increasing the value of $\tau_{CS}$ also increases the number of activated hyper-classes. This is expected, because according to the Eq. \eqref{eq:cs}, in order to meet the $\tau_{CS}$, a larger number of hyper-class classifiers should be activated. Fig. \ref{fig:metrics} captures the change in the accuracy and increase in the computational complexity (Flop count) when the $\tau_{CS}$ varies in that range. Figure \ref{fig:metrics}-top shows that increasing the $\tau_{CS}$ beyond 0.7 results in no gain in the accuracy. Increasing the $\tau_{CS}$ beyond 0.7 results in the activation of a larger number of hyper classifiers and an increase in computational complexity without any gain in accuracy. This implies that at this particular scenario the best $\tau_{CS}$ is 0.7.

\begin{table}[hbt!]

    \caption{Number of the activated hyper-classes when $\mathcal{T}_{CS}$ varies in the range 0.5 to 0.95. CC-ideal shows the hypothetical ideal scenario that only one hyper-class activated per each input sample. Noted the summation of activated hyper-classes at each row is 10000.  }
    \label{activated}

    \begin{adjustbox}{width=0.98\columnwidth,center}
    	\begin{tabular}{|c|c|c|c|c|c|c|c|c|c|}
    		\cline{1-7} \cline{9-10}
    		\cline{1-7} \cline{9-10}
    		\cellcolor{lightgray} $\mathbf{\tau_{CS}}$ & \cellcolor{lightgray} \textbf{\#1} & \cellcolor{lightgray} \textbf{\#2} & \cellcolor{lightgray} \textbf{\#3} & \cellcolor{lightgray} \textbf{\#4} & \cellcolor{lightgray} \textbf{\#5} & \cellcolor{lightgray} \textbf{\#6} &   & \cellcolor{lightgray} \textbf{Acc. Gain} & \cellcolor{lightgray} \textbf{CC. Save}\tabularnewline 
    		\cline{1-7} \cline{9-10} 
    		                    \textbf{0.5}	&9379	&619	&2		&0		&0	&0  &   &   -0.0638       & 0.3392\tabularnewline         
    		\cline{1-7} \cline{9-10} 
    		                    \textbf{0.6}	&8545	&1434	&21		&0		&0	 &0  &   &  -0.0274       & 0.2996\tabularnewline         
    		\cline{1-7} \cline{9-10} 
    		                     \textbf{0.7}	&7683	&2166	&150	&1		&0	 &0 &   &   0.0021        & 0.2575 \tabularnewline         
    		\cline{1-7} \cline{9-10} 
    		                     \textbf{0.8}	&6641	&2789	&546	&24		&0	 &0  &   &  0.0023         & 0.2000\tabularnewline         
    		\cline{1-7} \cline{9-10} 
    		                     \textbf{0.9}	&5198	&3294	&1228	&263	&17	 &0 &   &   0.0023         & 0.1137\tabularnewline         
    		\cline{1-7} \cline{9-10} 
    		                     \textbf{0.95} &4044	&3372	&1783	&684	&115 &2 &   &    0.0023        & 0.0368\tabularnewline         
    		\cline{1-7} \cline{9-10} 
    		                     \textbf{CC-ideal} &10000	&0	&0	&0	&0     &0    &   &  -0.2365       & 0.3781\tabularnewline         
    		\cline{1-7} \cline{9-10}

    	\end{tabular}
    \end{adjustbox}

\end{table}

Fig. \ref{fig:metrics}.(bottom) also captures the breakdown of the total computational complexity for different values of  $\tau_{CS}$ as it varies at the range (0.5, 0.95). Considering that in the evaluation set, we had an equal number of images from each class, it was expected that clusters with a higher number of member classes contribute to a lager FLOP count.

\section{Acknowledgement}
This work was supported by the National Science Foundation (NSF) through Computer Systems Research (CSR) program under NSF award number 1718538.

\section{Conclusion}
In this paper, we proposed conditional classification as an hierarchical CNN model that reaches a level of accuracy in the range of the state of the art solutions, with a significantly lower computational complexity. Our method uses a first stage CNN block ($\mathcal{S}$) to extract class independent features, utilizes a Mid-level Clustifier (($\mathcal{M}$) to identify the membership of the input image to one or few of the possible clusters, and then activates small and hyper-class specific classifier(s) to classify the input image. We illustrate how an existing model, such as ResNet 18, could be translated into our method. We reported up to 37\% reduction in overall computational complexity (compare to the original model) when ResNET was translated to its reduced counterpart with negligible loss in accuracy.  


\renewcommand{\IEEEbibitemsep}{0pt plus 0.5pt}
\makeatletter
\IEEEtriggercmd{\reset@font\normalfont\fontsize{6.3pt}{6.5pt}\selectfont}
\makeatother
\IEEEtriggeratref{1}

\bibliographystyle{IEEEtran}
\bibliography{main}

\begin{thebibliography}{10}
\providecommand{\url}[1]{#1}
\csname url@samestyle\endcsname
\providecommand{\newblock}{\relax}
\providecommand{\bibinfo}[2]{#2}
\providecommand{\BIBentrySTDinterwordspacing}{\spaceskip=0pt\relax}
\providecommand{\BIBentryALTinterwordstretchfactor}{4}
\providecommand{\BIBentryALTinterwordspacing}{\spaceskip=\fontdimen2\font plus
\BIBentryALTinterwordstretchfactor\fontdimen3\font minus
  \fontdimen4\font\relax}
\providecommand{\BIBforeignlanguage}[2]{{%
\expandafter\ifx\csname l@#1\endcsname\relax
\typeout{** WARNING: IEEEtran.bst: No hyphenation pattern has been}%
\typeout{** loaded for the language `#1'. Using the pattern for}%
\typeout{** the default language instead.}%
\else
\language=\csname l@#1\endcsname
\fi
#2}}
\providecommand{\BIBdecl}{\relax}
\BIBdecl

\bibitem{sayadi20192smart}
H.~Sayadi \emph{et~al.}, ``2smart: A two-stage machine learning-based approach
  for run-time specialized hardware-assisted malware detection,'' in \emph{2019
  Design, Automation \& Test in Europe Conference \& Exhibition (DATE)}.\hskip
  1em plus 0.5em minus 0.4em\relax IEEE, 2019, pp. 728--733.

\bibitem{vakil2020lasca}
A.~Vakil \emph{et~al.}, ``Lasca: Learning assisted side channel delay analysis
  for hardware trojan detection,'' \emph{arXiv preprint arXiv:2001.06476},
  2020.

\bibitem{azar2020nngsat}
K.~Z. Azar \emph{et~al.}, ``Nngsat: neural network guided sat attack on logic
  locked complex structures,'' in \emph{2020 IEEE/ACM International Conference
  On Computer Aided Design (ICCAD)}.\hskip 1em plus 0.5em minus 0.4em\relax
  IEEE, 2020, pp. 1--9.

\bibitem{vakil2020learning}
A.~Vakil \emph{et~al.}, ``Learning assisted side channel delay test for
  detection of recycled ics,'' \emph{Asia and South Pacific Design Automation
  Conf.}, 2021.

\bibitem{he2016deep}
K.~He \emph{et~al.}, ``Deep residual learning for image recognition,'' in
  \emph{Proceedings of the IEEE conference on computer vision and pattern
  recognition}, 2016, pp. 770--778.

\bibitem{sanders2010cuda}
J.~Sanders~et al., \emph{CUDA by Example: An Introduction to General-Purpose
  GPU Programming, Portable Documents}.\hskip 1em plus 0.5em minus 0.4em\relax
  Addison-Wesley Professional, 2010.

\bibitem{abadi2016tensorflow}
M.~Abadi \emph{et~al.}, ``Tensorflow: A system for large-scale machine
  learning,'' in \emph{12th $\{$USENIX$\}$ Symposium on Operating Systems
  Design and Implementation ($\{$OSDI$\}$ 16)}, 2016, pp. 265--283.

\bibitem{chen2014diannao}
T.~Chen \emph{et~al.}, ``Diannao: A small-footprint high-throughput accelerator
  for ubiquitous machine-learning,'' in \emph{ACM Sigplan Notices}, vol.~49,
  no.~4.\hskip 1em plus 0.5em minus 0.4em\relax ACM, 2014, pp. 269--284.

\bibitem{chen2014dadiannao}
Y.~Chen \emph{et~al.}, ``Dadiannao: A machine-learning supercomputer,'' in
  \emph{Proceedings of the 47th Annual IEEE/ACM International Symposium on
  Microarchitecture}.\hskip 1em plus 0.5em minus 0.4em\relax IEEE Computer
  Society, 2014, pp. 609--622.

\bibitem{du2015shidiannao}
Z.~Du \emph{et~al.}, ``Shidiannao: Shifting vision processing closer to the
  sensor,'' in \emph{ACM SIGARCH Computer Architecture News}, vol.~43,
  no.~3.\hskip 1em plus 0.5em minus 0.4em\relax ACM, 2015, pp. 92--104.

\bibitem{chen2016eyeriss}
Y.-H. Chen \emph{et~al.}, ``Eyeriss: A spatial architecture for
  energy-efficient dataflow for convolutional neural networks,'' in \emph{ACM
  SIGARCH Computer Architecture News}, vol.~44, no.~3.\hskip 1em plus 0.5em
  minus 0.4em\relax IEEE Press, 2016, pp. 367--379.

\bibitem{mirzaeian2019tcd}
A.~Mirzaeian \emph{et~al.}, ``{TCD-NPE: A Re-configurable and Efficient Neural
  Processing Engine, Powered by Novel Temporal-Carry-deferring MACs},''
  \emph{arXiv preprint arXiv:1910.06458}, 2019.

\bibitem{nesta}
A.~Mirzaeian~et al., ``Nesta: Hamming weight compression-based neural proc.
  engine,'' \emph{arXiv preprint arXiv:1910.00700}, 2019.

\bibitem{daneshtalab2020hardware}
M.~Daneshtalab~et al., \emph{Hardware Architectures for Deep Learning}.\hskip
  1em plus 0.5em minus 0.4em\relax Materials, Circuits and Device, 2020.

\bibitem{Faraji_ISCAS_2020}
S.~R. {Faraji} \emph{et~al.}, ``Hbucnna: Hybrid binary-unary convolutional
  neural network accelerator,'' in \emph{2020 IEEE International Symposium on
  Circuits and Systems (ISCAS)}, Oct 2020.

\bibitem{9116473}
A.~{Mehrabi} \emph{et~al.}, ``Prospector: Synthesizing efficient accelerators
  via statistical learning,'' in \emph{2020 Design, Automation Test in Europe
  Conference Exhibition (DATE)}, 2020, pp. 151--156.

\bibitem{10.1145/3427377}
\BIBentryALTinterwordspacing
A.~Mehrabi \emph{et~al.}, ``Bayesian optimization for efficient accelerator
  synthesis,'' \emph{ACM Trans. Archit. Code Optim.}, vol.~18, no.~1, Dec.
  2021. [Online]. Available: \url{https://doi.org/10.1145/3427377}
\BIBentrySTDinterwordspacing

\bibitem{lecun2015lenet}
Y.~LeCun \emph{et~al.}, ``Lenet-5, convolutional neural networks,'' \emph{URL:
  http://yann. lecun. com/exdb/lenet}, vol.~20, 2015.

\bibitem{neshatpour2018icnn}
K.~Neshatpour \emph{et~al.}, ``Icnn: An iterative implementation of
  convolutional neural networks to enable energy and computational complexity
  aware dynamic approximation,'' in \emph{2018 Design, Automation \& Test in
  Europe Conference \& Exhibition (DATE)}.\hskip 1em plus 0.5em minus
  0.4em\relax IEEE, 2018, pp. 551--556.

\bibitem{xiao2014error}
T.~Xiao \emph{et~al.}, ``Error-driven incremental learning in deep
  convolutional neural network for large-scale image classification,'' in
  \emph{Proceedings of the 22nd ACM international conference on
  Multimedia}.\hskip 1em plus 0.5em minus 0.4em\relax ACM, 2014, pp. 177--186.

\bibitem{Heidari-etal-2020-ELMO}
M.~Heidari~et al., ``Deep contextualized word embedding for text-based online
  user profiling to detect social bots on twitter,'' \emph{{IEEE} International
  Conference on Data Mining Workshops (ICDMW), {ICDMW}}, 2020.

\bibitem{srivastava2013discriminative}
N.~Srivastava~et al., ``Discriminative transfer learning with tree-based
  priors,'' in \emph{Advances in Neural Information Processing Systems}, 2013,
  pp. 2094--2102.

\bibitem{deng2014large}
J.~Deng \emph{et~al.}, ``Large-scale object classification using label relation
  graphs,'' in \emph{European conference on computer vision}.\hskip 1em plus
  0.5em minus 0.4em\relax Springer, 2014, pp. 48--64.

\bibitem{liu2013probabilistic}
B.~Liu \emph{et~al.}, ``Probabilistic label trees for efficient large scale
  image classification,'' in \emph{Proceedings of the IEEE conference on
  computer vision and pattern recognition}, 2013, pp. 843--850.

\bibitem{Heidari-etal-2020-Social_bots}
M.~Heidari~et al., ``Using bert to extract topic-independent sentiment features
  for social media bot detection,'' \emph{{IEEE} 11th Annual Ubiquitous
  Computing, Electronics \& Mobile Communication Conference, {UEMCON}}, 2020.

\bibitem{ng2002spectral}
A.~Y. Ng \emph{et~al.}, ``On spectral clustering: Analysis and an algorithm,''
  in \emph{Advances in neural information processing systems}, 2002, pp.
  849--856.

\bibitem{shi2000normalized}
J.~Shi~et al., ``Normalized cuts and image segmentation,'' \emph{Departmental
  Papers (CIS)}, p. 107, 2000.

\bibitem{panda2016conditional}
P.~Panda \emph{et~al.}, ``Conditional deep learning for energy-efficient and
  enhanced pattern recognition,'' in \emph{Design, Automation \& Test in Europe
  conf., 2016}.\hskip 1em plus 0.5em minus 0.4em\relax IEEE, 2016, pp.
  475--480.

\bibitem{teerapittayanon2016branchynet}
S.~Teerapittayanon \emph{et~al.}, ``Branchynet: Fast inference via early
  exiting from deep neural networks,'' in \emph{Pattern Recognition (ICPR),
  2016 23rd int. conf. on}.\hskip 1em plus 0.5em minus 0.4em\relax IEEE, 2016,
  pp. 2464--2469.

\bibitem{neshatpour2019icnn}
K.~Neshatpour and et~al., ``Icnn: The iterative convolutional neural network,''
  \emph{ACM Transactions on Embedded Computing Systems (TECS)}, vol.~18, no.~6,
  pp. 1--27, 2019.

\bibitem{neshatpour2018design}
K.~Neshatpour \emph{et~al.}, ``Design space exploration for hardware
  acceleration of machine learning applications in mapreduce,'' in \emph{2018
  IEEE 26th Annual International Symposium on Field-Programmable Custom
  Computing Machines (FCCM)}.\hskip 1em plus 0.5em minus 0.4em\relax IEEE,
  2018, pp. 221--221.

\bibitem{yan2015hd}
Z.~Yan \emph{et~al.}, ``Hd-cnn: hierarchical deep convolutional neural networks
  for large scale visual recognition,'' in \emph{Proceedings of the IEEE
  international conference on computer vision}, 2015, pp. 2740--2748.

\bibitem{von2007tutorial}
U.~Von~Luxburg, ``A tutorial on spectral clustering,'' \emph{Statistics and
  computing}, vol.~17, no.~4, pp. 395--416, 2007.

\end{thebibliography}

\end{document}